# A Fast Transformer-based General-Purpose Lossless Compressor

Yu Mao
Department of Computer Science,
City University of Hong Kong

Yufei Cui*
Department of Computer Science,
City University of Hong Kong

Tei-Wei Kuo
Department of Computer Science,
City University of Hong Kong
Department of Computer Science
and Information Engineering,
National Taiwan University

Chun Jason Xue
Department of Computer Science,
City University of Hong Kong

## ABSTRACT

Deep-learning-based compressor has received interests recently due to much improved compression ratio. However, modern approaches suffer from long execution time. To ease this problem, this paper targets on cutting down the execution time of deep-learning-based compressors. Building history-dependencies sequentially (e.g., recurrent neural networks) is responsible for long inference latency. Instead, we introduce transformer into deep learning compressors to build history-dependencies in parallel. However, existing transformer is too heavy in computation and incompatible to compression tasks.

This paper proposes a fast general-purpose lossless compressor, TRACE, by designing a compression-friendly structure based on a single-layer transformer. We first design a new metric to advise the selection part of compression model structures. Byte-grouping and Shared-ffn schemes are further proposed to fully utilize the capacity of the single-layer transformer. These features allow TRACE to achieve competitive compression ratio and a much faster speed. In addition, we further accelerate the compression procedure by designing a controller to reduce the parameter updating overhead. Experiments show that TRACE achieves an overall ~3x speedup while keeps a comparable compression ratio to the state-of-the-art compressors. The source code for TRACE and links to the datasets are available at https://github.com/mynotwo/A-Fast-Transformer-based-General-Purpose-LosslessCompressor.

## CCS CONCEPTS

- **Information systems** → **Data compression**.

## 1 INTRODUCTION

Data volume has been boosted in the past few years [5, 19]. Amazon Web Services (AWS) [40] contains more than 1.4 million hosts. These hosts transmit and store massive heterogeneous data streams every day including text, image, video, disk backup, MRI, to name a few. Large-scale data storage and transmission can be costly in data centers and cloud service providers. The German Climate Computing Center (DKRZ) [16] has a 54 PiB storage system that costs 0.9 GWh of energy per year, resulting in seven-figure dollars energy bill. This paper proposes a fast lossless deep-learning-based compressor to alleviate such burdens.

Standard compressors' performance are limited when facing multi-modal data streams [7–9, 14, 21, 25, 31, 35, 41]. One solution is to design a multi-specialized-compressor system to handle different data types [10, 11, 18, 20, 36–38]. Despite the effectiveness on specific data types, the scalability is often a major issue preventing it from being applied to the cloud. Another problem is data privacy since the distinguished compression ratio of specialized compressors is based on analyzing data characteristics.

Deep-learning-based general-purpose approaches have been recently studied as a better scheme in terms of high scalability [17, 22, 23, 28]. These compression methods can preserve data privacy while achieving general compression efficiency across data domains without knowing data type. Nevertheless, current deep-learning-based compressors are suffering from slow compression speed. Dictionary-based compressor [22, 23, 28] expands compression unit status, which can improve compression ratio but leads to drastically slow compression speed, usually among 20~150 kb/min. Byte-stream compressor sets compression unit as a byte, thus restricting compression unit status to $2^8$. Dzip [17] is a state-of-the-art byte-stream compressor that can improve compression speed to 399.4 kb/min.

This paper aims for a fast deep-learning-based general-purpose compression solution that improves compression speed, while maintaining comparable compression ratios across different data types. We replaced widely-applied *Recurrent Neural Network* (RNN) architecture with *transformer* as compressor's probability estimator, which alleviates poor GPU utilization caused by the sequential process in Dzip. However, standard transformer is model-heavy which leads to slow compression speed. In this paper, we first design a simple and effective metric, latency-to-compression ratio (LCR), to evaluate the efficiency and effectiveness of the compression model, taking both latency and compression ratio into account. We further investigate transformer's two main components: Attention and Feed-Forward Network (FFN)'s impact on compression. A latency-friendly baseline structure is selected under the guidance of LCR. After settling the baseline's structure, Byte-grouping and Shared-ffn are proposed to fully utilize a single-layer transformer's capacity. Byte-grouping decouples each byte's embedded dimension and Attention's hidden dimension to remove redundancy in feature vector space; Shared-ffn increases model capacity without additional parameters. These three features improve compression ratio compared with the un-modified single-layer transformer while maintaining fast execution speed. In addition, to mitigate the effects of frequent online adaptation, this paper proposes a Back-Prop (BP) Controller that discards useless parameter updates. This Back-Prop controller utilizes cross-entropy loss to automatically make local adaptation decisions with $O(1)$ complexity. This method leads to ∼ 30% speed up on average with 1% ∼ 2% compression ratio loss. The overall compression speed of the proposed approach is around three times faster than state-of-the-art while achieving a higher compression ratio across data domains.

---

*Corresponding Author.

## 2 MOTIVATION

Raising general-purpose compressors' compression ratio for multi-modal data can be challenging, because the byte stream is disordered, multi-distributional and without any prior knowledge. Standard compression algorithms such as Gzip [14] have a 3:1 compression ratio on Wikipedia data but only 1.37:1, 1.14:1, and 1.06:1 on sounds, images, and floating-point data, respectively.

To address this problem, deep-learning-based compressor research has gained a lot of traction recently [17, 22–24, 28]. NNCP [24] is a *transformer*-based compressor, which contains 55.16M params. In our evaluation, even with GPU acceleration, NNCP still takes 14 hours to compress a 95MB text file with one 2080Ti. This number increased to 60 hours when facing multi-modal disk backups. Cmix [22] is another highly optimized compressor but needs two days to compress a 95MB file in our experiments. This is because NNCP, as well as Cmix, are dictionary-based compressors. Statistically collecting patterns is effective, especially on text data. Unfortunately, dictionary-based method is the culprit of these compressor's slow compression speed because a large dictionary requires a large model to handle accurate prediction. On the other hand, compression gain by dictionary-based compressor drops sharply on non-text datasets.

Byte-stream compressors remove dictionary collection process and aim at precise probability estimation without any pre-processing. Although compression ratio could suffer on text data, byte-stream compression provides chances to accelerate general-purpose compression process by 10x with a much lighter probability estimator. Dzip [17] uses RNN as probability estimator and suffers from poor GPU memory utilization. RNN builds long-term dependency based on a sequential process, where each hidden state needs to wait for outputs from previous steps. Transformers build history dependency in parallel, which has better chance of fully utilize GPU.

Standard transformers have deep and heavy model structure whose parameters can often reach 300M [3, 15, 27, 36, 39]. This is due to the fact that transformer is originally designed to process linguistic knowledge. Therefore they need abilities to extract high-level semantic features from texts. Compression, though can also be modeled as sequence understanding task, focuses more on detecting low-level and repetitive patterns. We believe a single transformer layer is powerful enough but needs non-trivial modifications to adapt to compression tasks. This opinion is validated in experiment section.

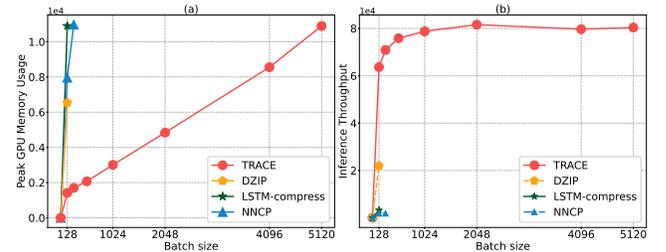

**Figure 1: (a) and (b) represents Peak GPU Memory Usage and Inference Throughput of RNN-based compressors and TRACE.**

Fig. 1 shows the proposed architecture's performance compared with standard compressor and RNN based compressors Dzip and Tensorflow-compress. For fairness, we remove Dzip's pre-training process since other methods all start compression from a random state. With batch size=128, Dzip and Tensorflow-compress's peak GPU memory usage are 6.39GB and 10.75GB. Such a high GPU memory usage limits their Inference Throughput to 21993 and 3213 bytes, respectively. As a comparison, the proposed method's GPU memory occupation is 1.4GB when batch size is 128, and the maximum throughput is 81594 bytes. This demonstrates the potential in the transformer-based compressor to achieve the goal of being fast and efficient at the same time.

## 3 PRELIMINARIES

### 3.1 Transformer Layer

The standard transformer layer includes two main sub-components: multi-head attention (MHA) and a fully connected feed-forward network (FFN).

**Multi-head attention (MHA).** There are three matrices in attention function: matrix $W^Q$, $W^K$, and $W^V$. The three matrices break input $x$ into three components: queries, keys, and values via linear projection denoted as $Q, K, V$. After the projection, queries and keys are multiplied to compute a measure of agreement, denoted as an attention matrix. Finally, the values matrix is multiplied by the attention matrix to produce a new representation for every input. The attention function can be formulated as follows:

$$\textbf{Attention}(x) = softmax(\frac{xW^Q * xW^{K^T}}{\sqrt{h_k}})xW^V, \quad (1)$$

where $h_k$ is the hidden dimension and acts as a scaling factor.

Instead of computing the attention only once, the multi-head mechanism runs through the scaled dot-product attention multiple times in parallel. The independent attention outputs are concatenated and linearly transformed into the standard dimensions.

$$MultiHead(Q, K, V) = [head_1; \ldots; head_h] W$$
$$\text{where } head_i = Attention\left(xW_i^Q, xW_i^K, xW_i^V\right), \quad (2)$$

where $W_i^Q, W_i^K, W_i^V$, and $W$ are parameter matrices to be learned.

**Feed-Forward Network (FFN).** Another important component of transformer layer is a fully connected feed-forward network, which can be formulated as follows:

$$FFN(x) = Gelu(xW_1)W_2, \quad (3)$$

where $W_1$ expand the dimension of MHA output and $W_2$ project activated high dimensional vector back to original vector space.

**Transformer Layer.** The full function of a transformer layer can be formulated as:

$$Transformer(x) = FFN(Attention(x)+x)+Attention(x)+x \quad (4)$$

## 3.2 Byte-Stream Data Compression

Byte-stream data compression involves two crucial steps: modeling and coding.

**Probability Estimation.** Any data can be represented as a binary data stream. Each 8 bit of this stream is considered as a byte to compress, whose value would fall into the range [0, 255]. The target of probability estimation model is to predict the probability distribution of the next byte using history bytes. Most of the deep-learning-based data compression works focus on this part, using history bytes as input and building models to extract correlations between history bytes and current byte.

**Coding.** The entropy theory introduced by Shannon gives out the lower bound of entropy coding, and there are many coders can get close to this boundary like Huffman Coding, Arithmetic Coding, etc. We use arithmetic coding in this work following the settings in [17, 23, 28].

**Whole compression procedure.** This paper focus on dynamic compression settings, which means the probability estimator start from a random state and dynamically adjust its parameters during compression procedure. The whole compression procedure can be expressed as follows:

---
**Algorithm 1: Dynamic Compression Process**

**Input:** Byte Stream$\{x_0, x_1, ..., x_{end}\}$, Context length $c$.
**Output:** Compressed file.
1 $i = 0$;
2 **while** $0 <= i < c$ **do**
3     Arithmetic Encode($x_i, \frac{1}{256}$);
4     $i+ = 1$;
5 **end**
6 Model.initialize();
7 **while** $c <= i < end$ **do**
8     $P(x_i|x_{i-c}, ..., x_{i-1})$ = Model.forward($x_{i-c}, .., x_{i-1}$);
9     Arithmetic Encode($x_i, P(x_i|x_{i-c}, ..., x_{i-1})$);
10    Model.backward($x_i, P(x_i|x_{i-c}, ..., x_{i-1})$, loss function);
11    $i+ = 1$;
12 **end**

---

## 4 THE TRACE FRAMEWORK

This section introduces proposed transformer structure and explains the design philosophy. We define context length as $c$, group number as $g$, and hidden dimension as $h$, the i-th byte is denoted as $x_i$. As shown in Fig. 2, to encode $x_i$, $c*g$ bytes before $x_i$ is performed byte grouping to get $g$ groups, each group is then projected to $h$-dim vector space to get context vectors. Furthermore, context vectors are fed into the proposed architecture to get estimated probability. Estimated probability is sent to Arithmetic Coder for encoding. At the same time, a copy of the estimated probability is sent to Back-prop controller to make backpropagation decisions. The same procedure is then conducted on $x_{i+1}, x_{i+2}, ..., x_{end}$.

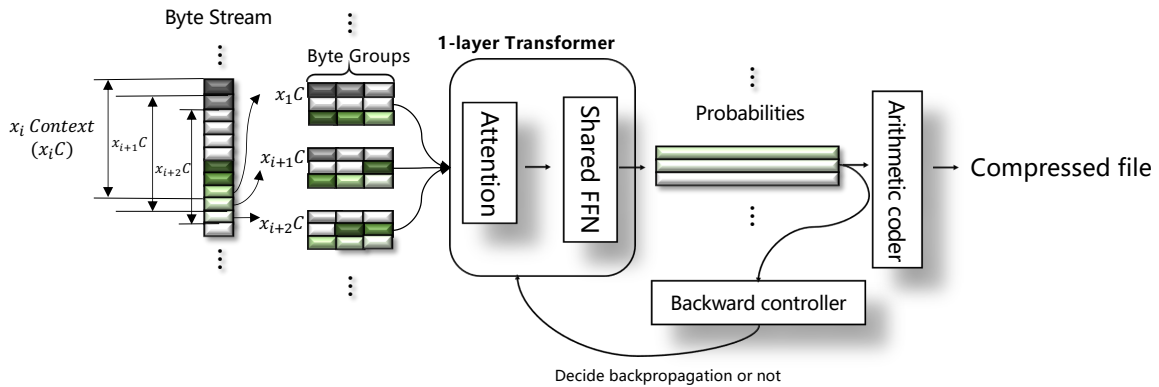

**Figure 2: Overall architecture of proposed TRACE.**

## 4.1 Latency-to-Compression Ratio (LCR)

Compression applications are aware of both compression latency and compression ratio. Improvements on these metrics can save financial cost and energy, especially when facing tremendous cloud data volumes. Unfortunately, a higher compression ratio comes with a greater model size, leading to higher latency. To the best of our knowledge, the previous metrics like compression speed, compression ratio, or bits-per-character only consider one aspect and cannot measure the trade-off between these two metrics.

Based on above consideration, we design a simple but effective metric *Latency-to-Compression Ratio (LCR)* to evaluate the run-time efficiency of compression model. Suppose $t_0$ and $t_i$ are inference latency for compression methods 0 and i. Meanwhile, $CR_0$ and $CR_i$ indicate the compression ratio of the two models. Definition of LCR would be:

$$LCR_i = \frac{t_i - t_0}{CR_i - CR_0} \quad (5)$$

Basically, LCR is the Latency increment with Compression Ratio improved by 1. Namely, if a model's compression ratio increases from 2 to 3, the LCR value indicates the latency increment. A larger LCR means faster latency growth of the the method, along with increasing compression ratio.

LCR is helpful to discover model's potential in workloads that are sensitive to the running speed. Moreover, to further prove LCR's potency, we conduct an empirical study on transformer architecture to explore how transformer components influence compression performance.

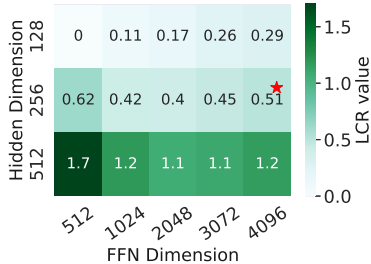

**Figure 3: Analysis of attention size (vertical) and FFN size (horizontal) with the value in the square to be the proposed LCR metric.**

The study is conducted on the Enwik8 dataset with a single-layer standard transformer model. Size (128, 512) is defined as a reference method $M_0$, where Attention dimension is 128 and FFN dimension is 512. We gradually increase the dimension of both modules and display corresponding LCR values in Fig. 3. Fig. 3 shows a clear trend that expansion on attention dimension always leads to a significantly larger LCR than FFN. This suggests that in terms of compression ratio, Attention is computational expensive while FFN is relatively cheap. Another phenomenon is that tinier models always have higher LCR, which indicates that as we increase model size, redundancies in the structure increases with effective part and lead to an increment in LCR. We further pick size (256, 4096) as a later experimental model setting since it has relatively low LCR with sufficient compression ratios.

## 4.2 Byte-grouping

In a standard NLP scenario, the token (the basic unit of NLP modeling) is projected into a high-dimensional vector space before being fed into a feature extractor. The dimension of the token vector space is bounded with the hidden dimension of the transformer, so the model can directly establish the relationship between tokens. In the data compression task, the token is replaced by bytes. However, a byte only has $2^8$ status while hidden dimension tends to be around $2^9$, $2^{10}$ or even larger. Hence, there would be redundancy in byte vector space if the byte vector dimension and hidden dimension are still bound together. This section introduces a method of decoupling byte vector dimension and hidden dimension, where adjacent bytes are combined to create a long-term dependence on the input bytes.

Conventional transformer use $\{x_{i-c}, ..., x_{i-1}\}$ to predict the probability of $x_i$, where $c$ is context length. Each byte is represented as a $v$-dim vector. In standard transformer $v$ is equal to the hidden dimension $h$, thus a byte can only calculate correlations with $c$ history bytes.

Adjacent bytes usually contain similar information thus can be grouped. Byte grouping procedure can be described as follows: First, we projected $c*g$ bytes into $h/g$-dim vector space. Any $g$ byte vector can form a group, so there are $c$ groups. Every group is concatenated together to create a group vector. Since every byte vector's dimension is $h/g$, each group will form a $h$-dim group vector. $c$ group vectors are then sent together into the attention module. In this way, a byte's context length can be extended $g$ times longer with no computation speed loss. Namely, the model can establish longer-range dependencies without sabotage in computational efficiency.

## 4.3 Shared Feed-Forward Network

Conventional transformers usually have 6 ∼ 12 layers to meet the needs of extracting high-level semantic features [3, 15, 27, 36, 39]. Nevertheless, data compression targets to predict probability of next byte and doesn't need such a high level feature extractor. To this end, this section first investigates the layer number's impact on compression ratio and compression speed, using a (256, 4096) transformer.

A Fast Transformer-based General-Purpose Lossless Compressor

Fig. 5 (a) represents 1-4 layers transformer's compression ratios on Enwik8. And Fig. 5 (b) shows compression speeds. After two layers, the increase in the compression ratio is negligible, which indicates transformer based compressors do not need as many layers as in NLP tasks. The two-layer model achieves a significantly larger compression ratio than one layer model. However, the compression speed is considerably worse, too. Shared-ffn is proposed to solve this trade-off between compression ratio and compression speed.

Generally, a transformer layer consists of an Attention followed by an FFN. The output is then transmitted to the next layer or a softmax layer to output. Analysis in Section.4.1 indicates that multiplying FFN number in a layer can be a potential direction to increase model capacity with less speed loss. However, deeper model means greater size. Inspired by ALBERT [26], which shared parameters between different layers, we propose a Shared-ffn which shares FFN parameters IN a layer. Shared-ffn would lead to a deeper model with a slight decrease in inference time. Moreover, this deeper layer is the same size as the general transformer layer. A 1-layer model under the (256, 4096) setting contains 2.4M parameters.

Fig. 4 (a) shows a typical transformer layer architecture. Input is first fed into attention module. The output of attention module is then transmitted to FFN. Fig. 4 (c) shows proposed technique, in which Shared-ffn replaces FFN. The output of Shared-ffn will return to the same FFN for $N$ times before feeding into next layer. $N$ can be as large as a user want without any parameter expansion. The experimental Shared-ffn depth is chosen as two.

Fig. 5 shows the compression ratio and compression speed of a 1-4 layer transformer and a 1-layer transformer with 2-Shared-ffn. "N-Shared-ffn" means the FFN is repeated by $N$ times. The result shows that Shared-ffn can further improve compression ratio on the basis of Byte-grouping while maintaining competitive compression speed.

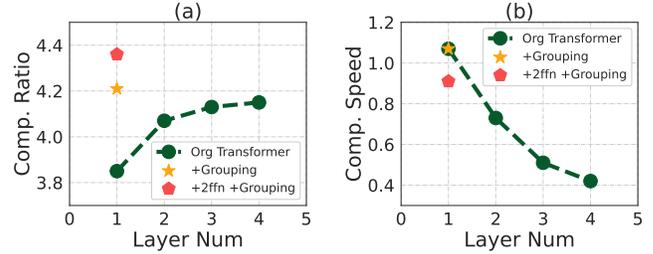

**Figure 5: Compression ratio and compression speed for 1-4 layer standard transformer, 1-layer transformer with Byte-grouping, and 1-layer transformer with 2 Shared-ffn and Byte-grouping on Enwik8.**

### 4.4 Dynamic compression with adaptive update scheme

Dynamic compression methods need back-propagation to maintain an update-to-date model. Local data distributions can be diverse. A simple text example is when the frequency of "apple" is higher than "amy" , model should give "p" a higher probability if "a" appears. As context changes, "amy" shows up more frequently than "apple", the model should give "m" a higher probability after "a". A more complicated case is heterogeneous data streams which contain a gallery of distributions. Thus, the model needs to adjust its prediction during compression.

Back-propagation provides an opportunity to fit model into local data distribution, but it also slows down the compression speed. Some compressors do back-propagation on each timestep during compression. Intuitively, data on a certain timestep sometimes contains similar information with history data, and it is unnecessary to compress with fully truncated backpropagation under such circumstances. We propose a BackProp (BP) Controller to detect timesteps with repetitive distribution and discard model updates on these timesteps.

We add a cache $LC$, which we represent by a tuple that stores recent $t$ cross-entropy value $e_i$. $e_i$ is calculated by one hot encoding $gt_i$ and estimated probability $P_i$ of bytes.

The cache has a fixed size $t$ and acts as a FIFO buffer: the eldest data is overwritten first when it is full. At each timestep, a dynamic threshold $T_i$ is retrieved from the cache with weights $w$. An incoming $e_i$ is compared with the retrieved threshold. If $e_i$ is larger, back-propagation is conducted for updating the parameters. Otherwise, we move to the next batch. The cache is updated with new entries, as they are seen, whether local adaptation is performed or not.

$$T_i = \frac{\sum_{n=i-t}^{i-1} w_n * e_n}{t} \qquad (6)$$

Because $e_i$ is already required by backpropagation, the only calculation we need to consider is the design of $w$. To maximize the benefit for compression speed bringing by the Back-Prop controller, we apply uniform weights in this case

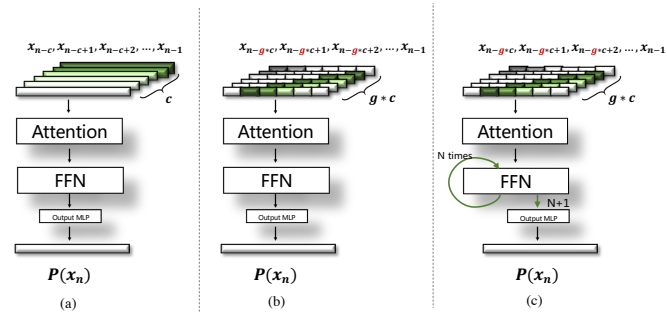

**Figure 4: Byte grouping and Shared-ffn illustration. (a) indicates the original architecture. (b) shows the architecture with byte grouping, where $x_i$'s context extends to $g*c$ history bytes. (c) shows the structure with Shared-ffn, where FFN repeated N times.**

**Algorithm 2: BackProp-Controller**

**Input:** FIFO Loss Cache $LC = \{e_{i-t}, e_{i-t+1}, ..., e_{i-1}\}$, Estimated Probabilities $P_i$, Symbols $X_i$

**Output:** Decision Policy 1 or Policy 2. **Policy 1**: Current data is "predictable" to model, don't backprop. **Policy 2**: Current data is "unpredictable" to model, backprop to update model parameters.

1 **foreach** *timestep i* **do**
2     Ground Truth $gt_i = OneHot(x_i)$;
3     Current Entropy $e_i = H(gt_i) + D_{KL}(gt_i \| P_i)$;
4     Threshold $T_i = Mean(e_{i-t}, e_{i-t+1}, ..., e_{i-1})$;
5     **if** $e_i >= T_i$ **then**
6         excute Policy 1;
7     **else**
8         excute Policy 2;
9     **end**
10     $LC.pop(e_{i-t})$;
11     $LC.append(e_i)$;
12 **end**

to avoid additional calculation. therefore Eq. 6 will degenerate into

$$T_i = Mean(e_{i-t}, e_{i-t+1}, ..., e_{i-1}) \quad (7)$$

This makes Back-Prop controller's complexity on each timesteps limits to $O(1)$. We also considered other approaches like clustering items in cache or calculate similarities. However, these methods would escalate complexity to at least $O(n)$, therefore, obey the target to accelerate the compression procedure.

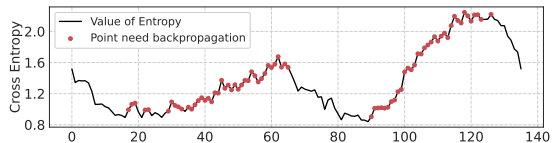

**Figure 6: Timsteps that BP controller decide to backprop on Enwik8.**

As shown in Fig. 6, this simple Back-Prop controller can detect the rise in entropy immediately and take actions to suppress the trend. As the entropy began to decrease, the controller stopped backpropagation to accelerate the compression procedure.

## 5 RELATED WORK

Ever since Shannon proposed the entropy coding method [34], some approaches, including statistical-based compressor, mixed context compressor, and deep learning-based compressor, have been made to achieve optimum compress ratio [7, 8, 18, 25, 37, 41].

There are some works designing compressors focused on specific types of data [10, 11, 18, 20, 36–38]. Costa et al. propose OUT [11] and SPATE [10], which can efficiently compress big telco data consists of network logs. AMMMO [1] is a two-level machine-learning-based compressor aiming at time-series data. BitSwap [20] generalized bits-back coding [37] and achieves better compression ratios in image lossless compression. IDF++ [38] discussed the potential difference in flexibility between invertible flows for discrete random variables and continuous random variables in the picture.

Another trend is constructing general-purpose compressors. A popular statistical-based method for lossless data compression is the Prediction by Partial Matching (PPM) [7], which based on a probability estimator and entropy coding that often takes arithmetic coding [41]. Another popular variant of PPM is PPM*C [8], which unbounded length contexts for PPM. PAQ [25] uses a context mixer same as PPM, and a mixed predictor, which combines the probability of different prediction models. CPU-based compressors such as Microsoft's Zipline [35] and Facebook's Zstd [9] offer compression ratios higher than standard deflate-based algorithms. CMIX [22] uses 2,122 mixed models including some neural networks and achieves highest compression ratio almost in all domains. However, CMIX is too slow to be applied in the real world.

More recently, there have been some attempts to use purely neural network-based architecture to fully utilize GPU's advantage in neural computing. They can be categorized into three classes below:

- Static-Pretrained-method: The model is pre-trained on file *F* for several epochs and freeze parameters during compression. This method is restricted to compress *F* and needs prolonged training time. Dzip-bootstrap [17] proposed a bootstrap model and DecMac [28] introduces a cycle connection to preserve hidden state ends while reusing them to initial following states. Some transformer variants' experiment on Enwik8 also fall into this setting.
- Dynamic-Pretrained-method: The model is pre-trained on *F* and still updates its parameter during compression. This method allows the model to dynamically fitting into the local context. However, adding backpropagation lead to slow speed. Dzip-combined [17] add another model upon Dzip-bootstrap and finetunes it during compression thus belongs to this category.
- Dynamic-Random-method: The model is random initialized and updates its parameter during compression. This method omits pre-training but may suffer from cold start problem. Knoll released Tensorflow-compress [23], which uses LSTM and CMIX's preprocessing. NNCP [2] is a transformer based compressor which add retrain procedure during compression. Proposed TRACE falls into this category.

## 6 EXPERIMENT

Experiments are benchmarked on several datasets from different domains. A wide variety of real-world datasets with different distributions are considered, including text (bookcorpus [39], enwik9 [29]), floating point data (obs-spitzer [4]), audio (environmental sound [32]), image data (ImageNet [12]), disk backup data and the Silesia [13]. The detailed description of those datasets can be found in Appendix.

We use Compression Ratio, Peak GPU Memory Usage, Compression Speed, Model Latency and Model Throughput as metrics. All experimental result is average of ten repetitive experiments. Model Latency and Model Throughput evaluate the computation efficiency of models including backpropagation time.

We compare TRACE with both standard compressors and deep-learning-based compressors. To ensure fairness, pre-training stage of Dzip is removed. Following the exploration in [36], a simplified Transformer architecture named Performer[27] is selected as the backbone. By default, all results corresponding to TRACE are obtained using an Attention module and a Shared-ffn. The hidden dimension and FFN dimension are set to 256 and 4096 respectively, following LCR's guidance in Section4.1. Byte group size, Shared-ffn layer number and context length are set to 4, 2 and 8. The compression batch size for TRACE is set to 512. Adam is used to optimize the model with learning rate=0.001. All results are reported on a 11GB NVIDIA Geforce RTX 2080 GPU. We also report Cmix, NNCP and Tensorflow's metrics including compression speed, peak GPU memory usage, inference latency and parameter amount, by smaller datasets (95 MB for each) that extracted from other workloads.

### 6.1 Computational Efficiency Comparison for Neural Network-based Compressors

This section presents the computational efficiency of Tensorflow-compress, NNCP, Dzip, and TRACE.

**Table 1: Peak gpu memory usage and Inference Latency of GPU-based method with batch=128, TC for tensorflow-compress.**

| Method | Peak GPU Memory Usage (GB) | Inference Latency (ms) | Params (M) |
|---|---|---|---|
| 2080Ti | 10.75G | - | - |
| TC | 10.64G | 39.84ms | 258.3M |
| NNCP | 7.75G | 95.67ms | 55.16M |
| Dzip | 6.39G | 5.82ms | 1M |
| TRACE | 2.02G | 1.46ms | 2.40M |

First we show the peak GPU memory usage, inference latency and model size of GPU-based compressors with batch size=128 in Table.1. The drawback of RNN-based compressors including Dzip and Tensorflow-compress reflect on their Peak GPU Memory Usage. Tensorflow-compress occupies nearly whole memory space while 1M Dzip model takes up half of the GPU memory on a typical single card 2080Ti, when batch size=128. NNCP's peak GPU memory usage is comparable to Dzip due to its transformer-based backbone, although the model is 55x larger and inference latency is 20x slower than Dzip. This demonstrates transformer's superiority in memory efficiency. Furthermore, TRACE only uses 1/3 of Dzip's GPU memory occupation even with a 2.4x larger model. The low memory occupation of TRACE indicates it can support a much larger batch than other compressors. Further reports for computation efficiency of TRACE under larger batch sizes are provided in Appendix.

**Table 2: Compression speed of Compressors.**

| Method | Compression Speed |
|---|---|
| Traditional | |
| Gzip | 1851MB/min |
| 7z | 1032MB/min |
| Zstd-19 | 191MB/min |
| Deep Learning-based | |
| Cmix | 39.9 kb/min |
| Tensorflow-compress | 30.72~163.8 kb/min |
| NNCP | 28.6~122.9 kb/min |
| Dzip | 399.4 kb/min |
| **TRACE** | **952.3** kb/min |
| **TRACE+BP controller** | **1228.8** kb/min |

Table. 2 shows the compression speed comparison between GPU-based compressors. We also report Cmix here since it's also based on Neural Networks.

An interesting phenomenon is that NNCP and tensorflow-compress's compression speed is unstable, changing with the data type. This is due to their dictionary-based compression process. To be specific, TensorFlow-compress and NNCP first collect repetitive patterns and obtain a dictionary. The original file is then interpreted to a shorter version using this dictionary. The following compression process is conducted on this short file (with a larger symbol space), and its length varies.

According to the NNCP's report [24], it spends 9 hours to compress 95MB[1] Enwik8 dataset on V100. Actually this is the speed of encoding of a 21MB projected file. This encoding time rises to 13.75h since our hardware is changed from V100 to 2080Ti, which is much slower. In the worst case of our selected datasets, the projected disk backup file expands to 84.78MB. Hence, the compression time of the disk backup file on 2080Ti increases to 55h, and the compression speed decreases to 0.028 MB/min.

Byte-stream compressors escape from this disaster. Dzip and TRACE both obtain stable, fast compression speed ignoring data stream type. But TRACE is 2x faster than Dzip. After equipping the Back-prop controller, TRACE's speed is further improved by 29%, which is 3x of Dzip.

---

[1] 95 MB is calculated from $10^8$ bytes

## 6.2 Ablation Study

We display the effectiveness of techniques introduced in Section 4. As shown in Table.3, with Byte-grouping and Shared-ffn, the model obtains an improved compression rate both on homogeneous data and heterogeneous data in all fields. Adding Back-prop controller will hurt compression ratios 1%~2% on all datasets while decrease on Silesia rises to 9.5%. This is because Silesia is 5x smaller than other datasets while containing various data types. Limited data amount would lead to insufficient training, which would confuse the Back-Prop controller initially. This phenomenon would alleviate larger heterogeneous datasets, which is proved in the experiment on the Backup dataset. Despite the dropping compression ratio (about 1%~2%), the Back-prop controller also provides 29% compression speed improvement.

## 6.3 Multi-Layer, Multi-Deep-ffn and Multi-Shared-ffn

Figure.7 shows the compression comparison of TRACE with 1-4 layers, 1-4 Deep-ffns (with seperate parameters) and 1-4 Shared-ffns (with shared parameters). Figure.7 (a) indicates the compression ratio on five datasets. It can be observed that increasing layer number performs almost the same as increasing the number of Shared-ffn on non-text data. As for two text datasets, Enwik9 and Book, more layers provide slightly more improvements than Shared-ffns in compression ratios. Deep-ffns' compression ratio is almost the same with multi-layer models.

Figure.7 (b) (c) (d) shows the metrics including Peak GPU Memory Usage, Latency and Throughput. Increasing layer number causes faster growth in GPU memory usage and estimation latency. A 4-layer model consumes 1.28x larger memory and is almost 2x slower than a 4-shared-ffn model. Shared-ffns run-time efficiency metric is almost the same with deep-ffns, and this is because pytorch's computational graph automatically copy shared-ffns nodes thus it would compute like deep-ffns, but with much fewer parameters.

## 6.4 Compression Performance on Large Datasets

Table.4 shows the proposed TRACE's compression ratio compared with the state-of-the-art general-purpose compressor Gzip, 7z, Zstd-19, and Dzip.

TRACE performs better in cross-domain compression tasks. On Enwik9 datasets, TRACE achieves a 5.29:1 compression ratio which is 94% higher than the best conventional method 7Z, and 82% higher than the byte-stream compression method Dzip. On another text data, BookCorpus, TRACE achieves a 4.58:1 compression ratio which is 1.8X higher than Gzip, 78% higher than 7Z, and 62% higher than Dzip.

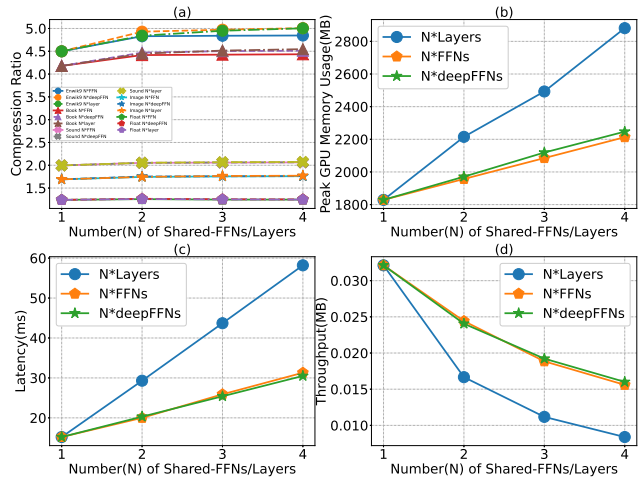

**Figure 7: Compression performance comparison of different layer/FFN number.**

On non-text data, the performance of traditional methods deteriorates rapidly, for traditional methods are based on statistical dictionaries, which is helpful for files with significant patterns such as text. However, it is hard to find repetitive sound, image, or floating-point files patterns. As for deep-learning-based approaches, Dzip is based on byte stream and gains better performance on this part. Dzip achieves compression ratios of 2.04:1, 1.72:1, and 1.26:1 on sound, image, and floating-point data, respectively. The proposed architecture achieves compression ratios of 2.16:1, 1.81:1, and 1.28:1 on average, which outperformed Dzip by 10%, 5%, and 1% respectively. However, TRACE performers are slightly worse than Dzip on 203M Silesia dataset because the transformer's complex structure leads to slow convergence.

As shown in Table. 4, TRACE performs better than unmodified transformer. Conventional transformer performs better than Dzip on BookCorpus, but worse on other datasets. Compared with conventional transformer, the compression ratio of TRACE on seven datasets are 23%, 8%, 11%, 9% and 2% higher, respectively. TRACE + BP Controller can maintain a comparable compression ratio with 29% improvement in compression speed on average.

**Table 3: Effectiveness validation of proposed techniques.**

| Methods | Compress Spd (kb/min) | Homogeneous Data | | | | | Heterogeneous Data | |
|---|---|---|---|---|---|---|---|---|
| | | Enwik9 | BookCorpus | Sound | Image | Floating Point | Silesia | Backup |
| Transformer (256, 4096) | 1064.9 | 4.31 | 4.26 | 1.94 | 1.66 | 1.26 | 4.03 | 1.61 |
| +Byte-Grouping (BG) | 1064.9 | 5.10 | 4.45 | 2.09 | 1.76 | 1.28 | 4.56 | 1.75 |
| +BG +Shared-ffn (SF) | 952.3 | 5.29 | 4.58 | 2.16 | 1.81 | 1.28 | 4.63 | 1.78 |
| +BG +SF +BP Control | 1228.8 | 5.12 | 4.49 | 2.13 | 1.79 | 1.27 | 4.23 | 1.75 |

Table 4: Compression Ratios on Large Datasets.

| Methods | | Homogeneous Data | | | | | Heterogeneous Data | |
|---|---|---|---|---|---|---|---|---|
| | | Enwik9 | BookCorpus | Sound | Image | Floating Point | Silesia | Backup |
| Traditional | Gzip | 3.09 | 2.77 | 1.37 | 1.14 | 1.06 | 3.10 | 1.28 |
| | 7z | 4.35 | 3.80 | 1.59 | 1.38 | 1.14 | 4.25 | 1.56 |
| | zstd-19 | 4.24 | 3.73 | 1.40 | 1.16 | 1.10 | 3.97 | 1.36 |
| Deep Learning Based | Dzip-combined | 4.47 | 3.95 | 2.04 | 1.72 | 1.26 | 4.78 | 1.78 |
| | Transformer (256-4096) | 4.31 | 4.26 | 1.94 | 1.66 | 1.26 | 4.03 | 1.61 |
| | TRACE | 5.29 | 4.58 | 2.16 | 1.81 | 1.28 | 4.63 | 1.78 |
| | TRACE+BP controller | 5.12 | 4.49 | 2.13 | 1.79 | 1.27 | 4.23 | 1.75 |

## 7 CONCLUSION

In this paper, we studied the under-explored computation efficiency problem in deep-learning-based general-purpose compressors. We proposed TRACE with computational-efficient designs of model structure to fully utilize the capacity of a single-layer transformer, which encompassed two levels of compression model latency reduction: reducing number of parameters and reducing byte vector space redundancy. We further accelerated the local adaptation procedure by avoiding useless backpropagation. A FIFO cache stores historical cross-entropy value that obtained to provide up-to-date thresholds for binary backward decisions. Our solution can maintain a high compression ratio and achieves a much faster compression speed. We demonstrated its superior efficiency and uncompromised quality on five homogeneous datasets and two heterogeneous datasets.

## REFERENCES

[1] X. Yu al. 2020. Two-Level Data Compression using Machine Learning in Time Series Database. In *2020 IEEE 36th International Conference on Data Engineering (ICDE)*. 1333–1344.
[2] Fabrice Bellard. 2019. NNCP: Lossless Data Compression with Neural Networks. (2019). https://bellard.org/nncp
[3] Nikolay Bogoychev. 2020. Not all parameters are born equal: Attention is mostly what you need. *arXiv preprint arXiv:2010.11859* (2020).
[4] M. Burtscher and P. Ratanaworabhan. 2009. FPC: A High-Speed Compressor for Double-Precision Floating-Point Data. *IEEE Trans. Comput.* 58, 1 (2009), 18–31.
[5] Worldwide Quarterly Enterprise Infrastructure Tracker: Buyer and Cloud Deployment. 2021. (2021).
[6] Kyunghyun Cho, Bart Van Merriënboer, Caglar Gulcehre, Dzmitry Bahdanau, Fethi Bougares, Holger Schwenk, and Yoshua Bengio. 2014. Learning phrase representations using RNN encoder-decoder for statistical machine translation. *arXiv preprint arXiv:1406.1078* (2014).
[7] J. Cleary and I. Witten. 1984. Data compression using adaptive coding and partial string matching. *IEEE Transacitons on Communications* 32, 4 (1984), 396–402.
[8] John G Cleary and William J Teahan. 1997. Unbounded length contexts for PPM. *Comput. J.* 40, 2_and_3 (1997), 67–75.
[9] Y. Collet. 2016. Zstd github repository from facebook. https://github.com/facebook/zstd
[10] C. Costa, G. Chatzimilioudis, D. Zeinalipour-Yazti, and M. F. Mokbel. 2017. Efficient Exploration of Telco Big Data with Compression and Decaying. In *2017 IEEE 33rd International Conference on Data Engineering (ICDE)*. 1332–1343.
[11] C. Costa, G. Chatzimilioudis, D. Zeinalipour-Yazti, and M. F. Mokbel. 2017. SPATE: Compacting and Exploring Telco Big Data. In *2017 IEEE 33rd International Conference on Data Engineering (ICDE)*. 1419–1420.
[12] J. Deng, W. Dong, R. Socher, et al. 2009. Imagenet: A large-scale hierarchical image database[C]//2009 IEEE conference on computer vision and pattern recognition. *Ieee* (2009), 248–255.
[13] Sebastian Deorowicz. 1985. Silesia Dataset. (1985). http://sun.aei.polsl.pl/sdeor/index.php?page=silesia
[14] Peter Deutsch. 1996. GZIP file format specification version 4.3. *RFC* 1952 (1996), 1–12. https://doi.org/10.17487/RFC1952
[15] J. Devlin, M. W. Chang, K. Lee, et al. 2018. Bert: Pre-training of deep bidirectional transformers for language understanding[J]. arXiv. (2018). arXiv:1810.04805 preprint.
[16] DKRZ. 2020. DKRZ. (2020). https://www.research-in-germany.org/en/research-landscape/research-organisations/research-infrastructures/dkrz.html
[17] Mohit Goyal, Kedar Tatwawadi, Shubham Chandak, and Idoia Ochoa. 2021. DZip: Improved general-purpose loss less compression based on novel neural network modeling. In *2021 Data Compression Conference (DCC)*. IEEE, 153–162.
[18] S. Idreos, R. Kaushik, V. Narasayya, and R. Ramamurthy. 2010. Estimating the compression fraction of an index using sampling. In *2010 IEEE 26th International Conference on Data Engineering (ICDE 2010*. 441–444.
[19] David Reinsel John Rydning, John F.Gantz. 2021. 2021–2025: The World Keeps Creating More Data — Now, What Do We Do with It All? (2021). https://www.idc.com/getdoc.jsp?containerId=US46410421
[20] Friso Kingma, Pieter Abbeel, and Jonathan Ho. 2019. Bit-swap: Recursive bits-back coding for lossless compression with hierarchical latent variables. In *International Conference on Machine Learning*. PMLR, 3408–3417.
[21] W Kinsner and RH Greenfield. 1991. The Lempel-Ziv-Welch (LZW) data compression algorithm for packet radio. In *[Proceedings] WESCANEX'91*. IEEE, 225–229.
[22] B. Knoll. 2014. CMIX. (2014). http://www.byronknoll.com/cmix.html
[23] B. Knoll. 2016. Tensorflow-compress. https://github.com/byronknoll/tensorflow-compress
[24] B. Knoll. 2020. NNCP: Lossless Data Compression with Neural Networks. (2020). https://bellard.org/nncp/
[25] Byron Knoll and Nando de Freitas. 2012. A machine learning perspective on predictive coding with PAQ8. In *2012 Data Compression Conference*. IEEE, 377–386.
[26] Zhenzhong Lan, Mingda Chen, Sebastian Goodman, Kevin Gimpel, Piyush Sharma, and Radu Soricut. 2019. Albert: A lite bert for self-supervised learning of language representations. *arXiv preprint arXiv:1909.11942* (2019).
[27] Valerii Likhosherstov, Krzysztof M Choromanski, Jared Quincy Davis, Xingyou Song, and Adrian Weller. 2021. Sub-linear memory: How to make performers slim. *Advances in Neural Information Processing Systems* 34 (2021).
[28] Qian Liu, Yiling Xu, and Zhu Li. 2019. DecMac: A Deep Context Model for High Efficiency Arithmetic Coding. In *2019 International Conference on Artificial Intelligence in Information and Communication (ICAIIC)*. IEEE, 438–443.
[29] Matt Mahoney. 2006. Large Text Compression Benchmark. (2006). http://mattmahoney.net/dc/text.html

Table 5: Description of Compression Datasets.

| Name | Size | Description |
|---|---|---|
| **Text Data** | | |
| Book | 954MB | First 1000M byte of BookCorpus [39]. |
| Enwik9 | 954MB | First 1000M byte of the English Wikipedia Corpus [29]. |
| Enwik8 | 95MB | First 10M byte of the English Wikipedia Corpus [29]. |
| **Floating Point Data** | | |
| Float | 1.1GB | Spitzer Space Telescope data showing a slight darkening [4]. |
| **Audio Data** | | |
| Audio | 842MB | ESC [32] Dataset for environmental sound classification. |
| **Image Data** | | |
| ImageNet | 1.2GB | 100000 pictures from ImageNet [12]. |
| **Heterogenous Data** | | |
| Backup | 954M | 1000M byte random extract from a disk backup. |
| Silesia | 206M | A compression benchmark dataset [13] that containing exe, image, database, pdf, etc. |

[30] José L Nunez-Yanez and Vassilios A Chouliaras. 2005. A configurable statistical lossless compression core based on variable order Markov modeling and arithmetic coding. *IEEE Trans. Comput.* 54, 11 (2005), 1345–1359.
[31] I. Pavlov. 1999. 7zip: File archiver. (1999). www.7-zip.org
[32] Karol J. Piczak. 2015. ESC: Dataset for Environmental Sound Classification. (2015). https://doi.org/10.7910/DVN/YDEPUT
[33] Jack W Rae, Anna Potapenko, Siddhant M Jayakumar, and Timothy P Lillicrap. 2019. Compressive transformers for long-range sequence modelling. *arXiv preprint arXiv:1911.05507* (2019).
[34] Claude Elwood Shannon. 1948. A mathematical theory of communication. *The Bell system technical journal* 27, 3 (1948), 379–423.
[35] Rajeev Sharma. 2016. Zipline. (2016). https://github.com/opencomputeproject/Project-Zipline
[36] Y. Tay, M. Dehghani, S. Abnar, et al. 2020. Long Range Arena: A Benchmark for Efficient Transformers. (2020). arXiv:2011.04006 preprint.
[37] James Townsend, Tom Bird, and David Barber. 2019. Practical lossless compression with latent variables using bits back coding. *arXiv preprint arXiv:1901.04866* (2019).
[38] Rianne van den Berg, Alexey A Gritsenko, Mostafa Dehghani, Casper Kaae Sønderby, and Tim Salimans. 2020. Idf++: Analyzing and improving integer discrete flows for lossless compression. In *International Conference on Learning Representations*.
[39] Ashish Vaswani, Noam Shazeer, Niki Parmar, Jakob Uszkoreit, Llion Jones, Aidan N Gomez, Łukasz Kaiser, and Illia Polosukhin. 2017. Attention is all you need. *Advances in neural information processing systems* 30 (2017).
[40] Wikipedia. 2020. Amazon Web Serivce. (2020). https://en.wikipedia.org/wiki/Amazon_Web_Services
[41] Ian H Witten, Radford M Neal, and John G Cleary. 1987. Arithmetic coding for data compression. *Commun. ACM* 30, 6 (1987), 520–540.
[42] Jacob Ziv and Abraham Lempel. 1977. A universal algorithm for sequential data compression. *IEEE Transactions on information theory* 23, 3 (1977), 337–343.

## A DATASET DESCRIPTIONS

Table.5 introduces the datasets for compressin in details.

## B SELECTION OF GROUP NUMBER AND CONTEXT LENGTH

This section reports compression performance and provides the selection basis of TRACE's different architecture settings, including group number and context length.

The impact of group number on compression ratio is reported in Fig.8 (a). As the hidden dimension is settled as 256, we chose 2,4,6,8 as experimental group numbers. Group number 6 on Enwik9 and Bookcorpus is better than other group sizes. In terms of Sound and Image data, group number 8 and group number 2 have a slightly higher compression ratio than other group sizes. However, the difference is not as apparent as in text data. In the case of float-point data, other group sizes achieve the same compression rate except group number 4. In general, setting group size to 4 can achieve a relatively high compression rate for all data. That is, a 64-dimensional vector represents each byte (256/4).

## C COMPRESSION PERFORMANCE OF CMIX, NNCP, DZIP AND TRACE

Many existing RNN-based or transformer-based compressor requires a long time to execute. To compare with those methods, we extract the first $10^8$ Bytes (95MB) for each dataset in the previous section as subsets to perform experiments in this section.

As shown in Table. 6, compared with Cmix and NNCP, the compression ratio of Dzip and TRACE in this part is worse on text data. This is because Cmix and NNCP are dictionary-based compressors, which means they would pass the whole file first to extract repeated patterns, obtain a dictionary, and project the original file to a smaller file. The technique is helpful for text data because projection does the main compression job. Unfortunately, this technique is the main culprit for their retard compression speed, not to mention the sharp improvement of dictionary drops on non-text datasets. Hence, byte-stream compressors like Dzip and TRACE remove the dictionary collection process and aim at precise probability estimation without pre-processing. Although compression ratio could suffer on text data, byte-stream compression provides chances to accelerate the general-purpose compression process by 10x with a much lighter probability estimator. We leave the improvement on text data as future works.

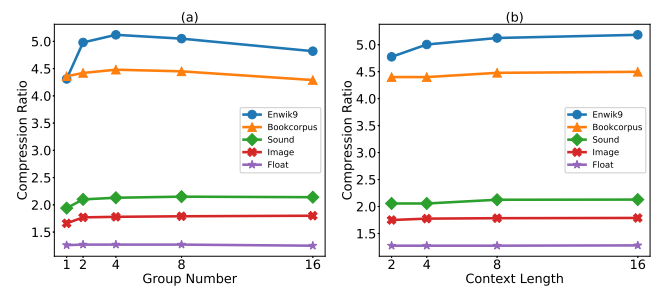

Figure 8: Compression Ratio of different group size.

Table 6: Compression Performance of Cmix, NNCP, Dzip and TRACE on subsets.

| Methods | Compression Speed | Homogeneous Data | | | | | Heterogeneous Data | |
|---|---|---|---|---|---|---|---|---|
| | | Enwik9 | BookCorpus | Sound | Image | Floating Point | Silesia | Backup |
| Cmix | 39.9 kb/min | 6.49 | 5.90 | 2.46 | 2.46 | 1.39 | 6.65 | 1.97 |
| NNCP | 28.6~122.9 kb/min | 5.85 | 5.57 | 2.09 | 1.76 | 1.26 | 5.64 | 1.75 |
| Dzip-combined | 399.4 kb/min | 4.35 | 4.27 | 1.99 | 1.71 | 1.24 | 4.78 | 1.70 |
| TRACE | 952.3 kb/min | 4.36 | 4.30 | 2.03 | 1.76 | 1.25 | 4.63 | 1.75 |

## D THROUGHPUT OF TRACE

Table 7: Peak gpu memory usage and Inference Latency of Different Batch Size.

| Method-Batch | Peak GPU Memory Usage (GB) | Inference Latency (ms) | Inference Throughput (byte/s) |
|---|---|---|---|
| Standard 2080Ti | 10.75 | - | - |
| TRACE-128 | 1.38 | 2.01 | 63682 |
| TRACE-256 | 1.66 | 3.61 | 70915 |
| TRACE-512 | 2.02 | 6.75 | 75852 |
| TRACE-1024 | 2.94 | 13.00 | 78769 |
| TRACE-2048 | 4.73 | 25.10 | 81594 |
| TRACE-4096 | 8.35 | 51.40 | 79688 |
| TRACE-5120 | 10.14 | 63.71 | 80364 |

Table. 7 shows TRACE's peak GPU memory usage, inference latency and inference throughput changes with batch size. The batch size is increased exponentially starting from batch size=128. TRACE can achieve batch size=5120 in the end while occupying 10.64G memory on a single 11G 2080Ti.